\documentclass[preprint,12pt,authoryear]{elsarticle}

\usepackage{amsmath}
\usepackage{unicode-math}

\usepackage{booktabs}
\usepackage{csquotes}
\usepackage{hyperref}
\usepackage{siunitx}
	\DeclareSIUnit{\quantity}{\relax}
	\DeclareSIUnit{\words}{words}
	\DeclareSIUnit{\sentences}{sentences}
\usepackage{zref-clever}
	\zcsetup{
		cap=true,
		abbrev=false
	}
\usepackage{xurl}
\usepackage{zref-titleref}

\journal{Journal of Language Modeling}

\begin{document}

\begin{frontmatter}

\title{surprisal is Not a Theory} %

\author{Andrés Buxó-Lugo\fnref{ub}}
\author{Aniello De Santo\fnref{utah}}
\author{Morgan Grobol\fnref{paris}}
\author{Ryan J. Hubbard\fnref{albany}}
\author{Cassandra L. Jacobs\corref{cor4}\fnref{ub}}
\ead{cxjacobs@buffalo.edu}

\affiliation[ub]{organization={University at Buffalo},
            city={Buffalo},
            state={NY},
            country={USA}}
\affiliation[utah]{organization={University of Utah},
            city={Salt Lake City},
            state={UT},
            country={USA}}
\affiliation[paris]{organization={Université Paris Nanterre},
            city={Paris},
            country={France}}
\affiliation[albany]{organization={University at Albany},
            city={Albany},
            state={NY},
            country={USA}}

\cortext[cor4]{Corresponding author}
\cortext[equal]{All authors contributed equally.}

\begin{abstract}
Surprisal Theory is often characterized as a \emph{computational-level} explanation per
\citep{marr_vision_1982}. 
We argue in this work that, even though a computational-level
narrative has been used to support \enquote{representation-agnostic} research within computational psycholinguistics, the movement toward black box systems embodied by large language models (LLMs)
does not exempt modelers using the surprisal metric from the representational decisions required by computational-level
characterizations. 
In fact, we argue that the uncritical use of LLM-surprisal obfuscates the representational and algorithmic-level commitments of different models.
In three analyses, we show that the choice of algorithm and model architecture plays a significant role in the computation of language model probabilities. We advise that researchers who wish to test Surprisal Theory re-evaluate the practice of treating large language model probabilities as interchangeable.
\end{abstract}

\begin{keyword}
surprisal \sep multiple realizability \sep theory-building

\end{keyword}

\end{frontmatter}

\section{Introduction}

The extent to which \emph{predictability} plays a fundamental role in human sentence processing is one of the central questions of modern (computational) psycholinguistics.
In this respect, \citet{hale_probabilistic_2001} pioneered the use of statistical parsing algorithms as models of human incremental sentence processing.
Hale proposed that a monotonic relationship should exist
between the probabilities assigned to structural options for incomplete sentences by a probabilistic parser and reading times in word-by-word self-paced reading.
This approach was further refined by
\citet{levy_expectation-based_2008}, who leveraged empirical methods to estimate syntactic probabilities from treebank corpora to support a model of processing complexity as a resource
allocation problem.
These two papers formed the foundation for \emph{Surprisal Theory}, which broadly applies \citeauthor{shannon_mathematical_1948}'s \citeyearpar{shannon_mathematical_1948} Information Theory framework to human language use.

At its core, Surprisal Theory aims to offer a direct linking hypothesis between cognitive effort during
language processing and properties of the linguistic input.
Concretely, this linking hypothesis takes the form of a complexity metric ---surprisal\footnote{Through the paper, we follow \citet{staub_predictability_2025} and \citet{slaats_whats_2025} in using upper-case Surprisal to refer to the conceptual, theoretical framework of Surprisal Theory, and lower-case surprisal to refer to the specific
operationalized metric.} --- operationalized by calculating the negative log probability of
an event -- for instance, observing a word given N preceding words.
This metric can be understood as a
measure of how \enquote{expected} a linguistic event was, and therefore how\enquote{effortful} it was to process.
Intuitively, if a word is very unexpected in a sentence, then surprisal is high, and reading times will be longer.
\citet{futrell_lossycontext_2020} helpfully formulate this claim as follows:
\begin{equation}
	D_{\mathrm{surprisal}}(w_i \mid c)\propto -\log p(w_i \mid c)
\end{equation}
where \(D\) stands for the difficulty of a word \(w_i\) in a particular context \(c\).

Since the formulation of Surprisal Theory \citep{hale_probabilistic_2001}, researchers have evaluated its validity primarily by correlating surprisal with measures of behavioral phenomena that are of interest to psycholinguistic researchers, and are believed to reflect cognitive effort (e.g., reading times).
Through these efforts, many papers have claimed that the empirical evidence supports Surprisal Theory as an omnibus account of processing difficulty \citep{michaelov_strong_2024,shain_large-scale_2024,smith_effect_2013}, though see a.o.\ \citet{huang_large-scale_2024} and \citet{slaats_whats_2025}.
Given this apparent empirical success of Surprisal Theory, it is worth revisiting how it has advanced our understanding of the phenomenon under study: the mechanisms underlying sentence processing.

Initial formulations of Surprisal Theory made explicit commitments to explanatory components of the language processing system \citep{hale_probabilistic_2001,hale_informationtheoretical_2016}.
This earlier work paid particular attention to the kind of probabilities that were being estimated, and how the process of obtaining them (e.g., from n-grams, context-free grammars, etc) was informed by the representational assumptions of different linguistic theories.
More recently however, the term \enquote{language model} has predominantly shifted from referring to models built on n-gram statistics \citep{smith_effect_2013} and parser probabilities to contemporary neural network-based (large) language models \citetext{LLMs; \citealp{bengio_neural_2003}}, and has similarly expanded the range of models used to estimate probabilities in psycholinguistics.
Currently, Surprisal Theory is often essentially equated with (transformations of) probabilities extracted from LLMs as measures of cognitive effort. Indeed, our understanding of the relationship between processing speed and probabilistic processing has been increasingly argued to be possible to observe because of the power of larger, neural language models to estimate surprisal \citep{goodkind_predictive_2018,shain_large-scale_2024}.

Simultaneous with this change in methodology, the field has moved away from trying to understand
measurable effects caused by experimental manipulations as a window into the function of complex
cognitive capacities, and toward \enquote{fitting} models of such effects
\citep{cummins_how_2000,van_rooij_theory_2020,van_rooij_theory_2021}.
That is, much recent work has ignored the search for explanatory processes in favor of developing increasingly powerful predictors
of psychophysical measures (i.e., predictive power, or \emph{PP};
\citealp{goodkind_predictive_2018,kuribayashi_lower_2021,wilcox_language_2023}).\footnote{We are not alone in our observation that the increasing technical focus on PP has limited the field's~ability to generate insights into deep theoretical questions about neural language models and psycholinguistic theory (see e.g., \citet{castro_formalism-implementation_2025} for a highly similar development in reinforcement learning theory).} 
It is now common to see research articles compute surprisal measure from dozens to hundreds of language models and, assuming that they are interchangeable, use them to study whether a word’s predictability in context reflects processing difficulty
\citep{shain_large-scale_2024,timkey_eye_2025,wilcox_predictive_2020,wilcox_targeted_2021}.

The practice of treating the predictive power of different LLMs are interchangeable is sometimes justified by the belief that \enquote{representation-agnosticism} \citep{futrell_lossycontext_2020} is a desirable consequence of using LLMs to estimate surprisal in accounts of psycholinguistic behavior. \citet{futrell_lossycontext_2020} summarize this claim as:

\begin{displayquote}[\citealp{futrell_lossycontext_2020}][.]
	In addition to providing an intuitive information-theoretic and Bayesian view of language
	processing, surprisal theory has the theoretical advantage of being representation-agnostic: The
	surprisal of a word in its context gives the amount of work required to update a probability
	distribution over any latent structures that must be inferred from an utterance. These
	structures could be syntactic parse trees, semantic parses, data structures representing
	discourse variables, or distributed vector-space representations. This
	representation-agnosticism is possible because the ultimate form of the processing cost function
	\textelp{} depends only on the word and its context, and the latent representation literally
	does not enter into the equation
\end{displayquote}

It is tempting, under a fitting-centric modeling paradigm, to infer that surprisal is derived directly from words and abstracted from intermediate representations, such as syntax.
However, we see representation-agnosticism as directly related to the problem of multiple realizability \citep{fodor_special_1974,putnam_psychological_1967}.
In reality, the surprisal measures that are calculated by different end-to-end black box language models rely on substantially different latent structures that are strongly conditioned by model design, and these latent structures in turn condition the information that is implicitly used to compute surprisal.
In other words, while LLM-based surprisal can be superficially similar across models, each model is algorithmically distinct and uses different types of information present in the input.
Consequently, this has led many researchers to engage in a practice of searching for the best-fitting language models through model comparison \citep{shain_large-scale_2024,oh-2026-model}

While it might be possible to compute estimates of surprisal without having to pay \emph{explicit attention} to representations, assumptions about representations are baked into the computation of the metric and should thus \emph{be an explicit concern} to computational modelers invested in explanatory theory building.
Accordingly,  the assumption that surprisal estimates from different models are interchangeably encoding the same abstract linking hypothesis is only theoretically valid if the metric is actually computed in \emph{functionally equivalent ways} across models.
Functional equivalence allows for incidental properties like differences in training data but requires that all subsequent processing remain identical \citep{rogers_changing_2021}.
Additionally, the latent representations should alter neither the \enquote{shape} of the cost function nor what that function encodes \citep{futrell_lossycontext_2020}.

In this paper, we argue that these assumptions are is in fact not wholly valid and are a consequence of a naive computationalist lens \citep{guest_logical_2023,guest_metatheory_2025}.
We also illustrate that the \enquote{black box} problem combined with outsourcing the computation of surprisal to LLMs diminishes the explanatory potential of Surprisal Theory, since different LLMs instantiate different implementations of the linking hypothesis between probabilities and effort.
We explore these issues with a series of experiments probing the inner workings of popular models used in the literature, which casts doubt on the viability of representation-agnosticism. 
We then present a critique of current trends in computational psycholinguistics informed by an understanding of \citeauthor{marr_vision_1982}'s \citeyearpar{marr_vision_1982} levels of explanation, and of the goals of computational cognitive models \citep{embick_towards_2015,guest_how_2021}.

Finally, we examine claims that (a) Surprisal Theory as encoded by surprisal is a computational-level description of the language comprehension process
\citetext{\citealp{hale_informationtheoretical_2016,futrell_noisy-context_2017}; though see
\citealp{hoover_plausibility_2023,schuler_evaluation_2024}}, and (b) that the successes of LLM-surprisal metrics have been informative in advancing theories of sentence processing.
In light of these considerations, we propose that researchers who wish to advance Surprisal Theory as an account of sentence processing must move beyond the dominant data-fitting paradigm, and explicitly evaluate the assumptions underlying the models they use to operationalize surprisal estimates. 

\section*{Experiments}

In \ztitleref{sec:expe_lexical_predictability}, we show that while predictions from different LLMs
correlate well with human behavior in one psycholinguistic task, these predictions evolve across
layers in very different ways, and these strong correlations are primarily driven by
higher-probability observations.
In \ztitleref{sec:expe_lexicality}, we demonstrate through an
analysis of the LLMs' hidden states that the types of representations that LLMs produce during the
prediction process are starkly different.
The results of both experiments provide evidence that
different language models encode lexical information in different ways across their hidden states.
We discuss these findings with respect to representation-agnosticism in Surprisal Theory and Computational-level accounts of language processing.
We include data and visualization scripts from all Experiments on the Open Science Foundation project page (\url{https://osf.io/n8kb3/overview}).

\section[Experiment 1]{Experiment 1: Evolution of lexical predictability across layers}\label{sec:expe_lexical_predictability}

The goal of \ztitleref{sec:expe_lexical_predictability} is to assess how language models assign
probabilities to future outcomes across layers by comparing their behavior to humans in a similar
task. For the human data, we conceptualize predictability using cloze probabilities, which measure
for a particular sentence context the proportion of human guesses that correspond to a particular
word. The shift among psycholinguists to focus on incremental (word-by-word) processing in tasks
such as self-paced reading\citep{aaronson_performance_1976} and rapid serial visual presentation
\citep{kieras_new_1984} has led to a growth in cloze corpora
\citep{de_varda_cloze_2023,jacobs_uncovering_2025,luke_limits_2016,peelle_completion_2020} that can
provide a lens through which we can interpret language model probabilities. We stratify human cloze
responses along their cloze probabilities in order to probe the layerwise computation of language
model probabilities for sentence-final completions.

\subsection{Data}

We focus our analyses here on \citeauthor{peelle_completion_2020}'s
\citeyearpar{peelle_completion_2020} cloze norms. This dataset contains \num{3085} sentences and
approximately \num{52000} unique, mostly sentence-final, cloze responses that vary in their
sentential constraint, or the degree to which the final word could be predicted by the upstream
context. That is, the sentences were designed to elicit varying degrees of agreement about the identity of the most likely completion of that sentence. The dataset includes
cloze probabilities for each preamble and other production statistics, which are aggregated across participants as a proportion of a word given the preceding context. We include sample sentences and their top completions in \zcref{tab:peele_cloze}.

\begin{table}[t]
	\centering
	\caption{Example preambles and human completions from \citeauthor{peelle_completion_2020}'s
\citeyearpar{peelle_completion_2020} cloze norms.}\label{tab:peele_cloze}
	\bigskip
     \resizebox{\textwidth}{!}{%
		\begin{tabular}{lcccc}
			\toprule
			Preamble & Ans 1 & Ans 2 & Ans 3 & Ans 4 \\
			\midrule
			He hated bees and feared encountering a & hive & swarm & bee & nest \\
			The baby's face puckered when she ate something & sour &
			salty & bitter & slimy \\
			Hearing noises from above, the confused man inspected the & attic &
			ceiling & roof & sky \\
			\bottomrule
		\end{tabular}
	}	
\end{table}

\subsection{Models}\label{sec:models}

We examine three Transformer language models commonly used in the Surprisal Theory literature that are roughly equivalent in the number of parameters that they use to solve linguistic prediction tasks. Specifically, we select models that are similar in size to GPT-2, due to reports that this model produces the strongest correlations with measures of sentence processing effort \citep{shain_large-scale_2024, oh_leading_2024, nair_words_2023}. 
Some work has indicated that there are also diminishing returns with respect to the strength of correlation that can be obtained in psycholinguistics, such that larger models may fail to produce appropriate predictability estimates \citep{10.1162/tacl_a_00548}, and recent literature has come to use these systems as a baseline.
We rely on two decoder-only models, Pythia-160M
\citep{biderman_pythia_2023} and GPT-2 (small size; \citealp{radford_language_2019}), which perform the task of next-word prediction, taking a sequence of words as input, and outputting a probability distribution over the possible values for the next token in the sentence.
For these models, we can easily compute next-word probabilities by assessing the predictions at each token.
The third model, RoBERTa (\enquote{base} size; \citealp{liu_roberta_2019}), is an encoder-only Transformer model, and is trained as a \emph{masked language model} that produces probabilities for masked tokens at arbitrary positions in a complete sentence, taking into account both the left- and right-context. 
For RoBERTa, we obtain next-word probabilities given a partial sentence by appending a mask token to it and probing the model's guesses at that token.
This procedure slightly deviates from the training regime.

Beyond the training objectives that distinguish the encoder and decoder model families, another interesting architectural difference across the models is the nature of the input and output transformations. 
The unembedding matrix \(U\) is responsible for transforming the compressed hidden state representation of the context into probability distribution over tokens; the embedding matrix \(E\) transforms tokens into hidden state representations.
GPT-2 and RoBERTa require, as an architectural constraint, that the embedding matrix that produces the initial embeddings of tokens be to the transpose of the unembedding matrix \(U\) used by their language modeling heads to transform the output of their last hidden layer into a score distribution over their vocabulary, a technique known as \emph{tied embeddings} \citep{press_using_2017}. 
Pythia, on the other hand, is not trained with this requirement, and therefore separates the embedding and unembedding matrices.

In this experiment, we conduct two probing studies. To assess the relationship between human cloze
and language model probabilities, and particularly the computation of next-word prediction, we apply
the \enquote{logit lens} technique \citep{dar_analyzing_2023}, which produces a distribution of
activations of potential states. We first examine how the language model logits correlate to human
cloze probabilities across layers (\ztitleref{sec:study_layerwise_correlation}) Then, we examine the
logits themselves across the different models, targeting words of differing levels of predictability
(\ztitleref{sec:sudy_layerwise_encoding}).

\subsection[Study 1.1]{Study 1.1: Layerwise correlation of cloze probabilities to language model
probabilities}\label{sec:study_layerwise_correlation}

\begin{figure}[t]
	\centering
	\includegraphics[width=\textwidth]{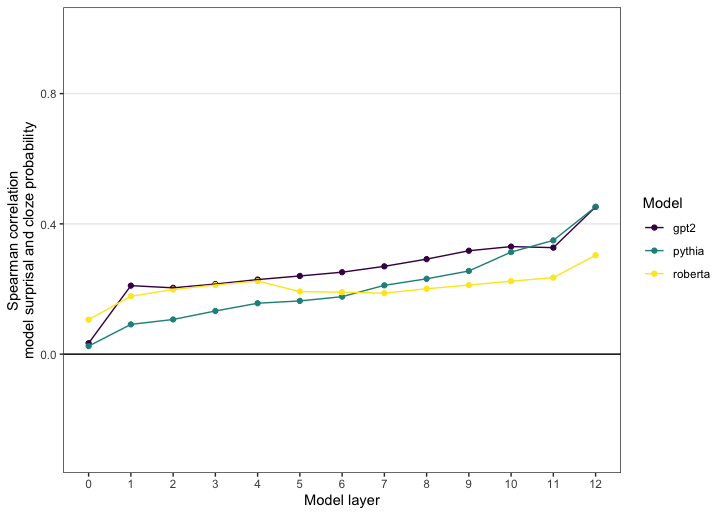}
	\caption{Spearman's correlation between word model-estimated surprisal and cloze probability, across models and layers.}\label{fig:layerwise_clozeprob_correlations}
\end{figure}

\begin{figure}[t]
	\centering
	\includegraphics[width=\textwidth]{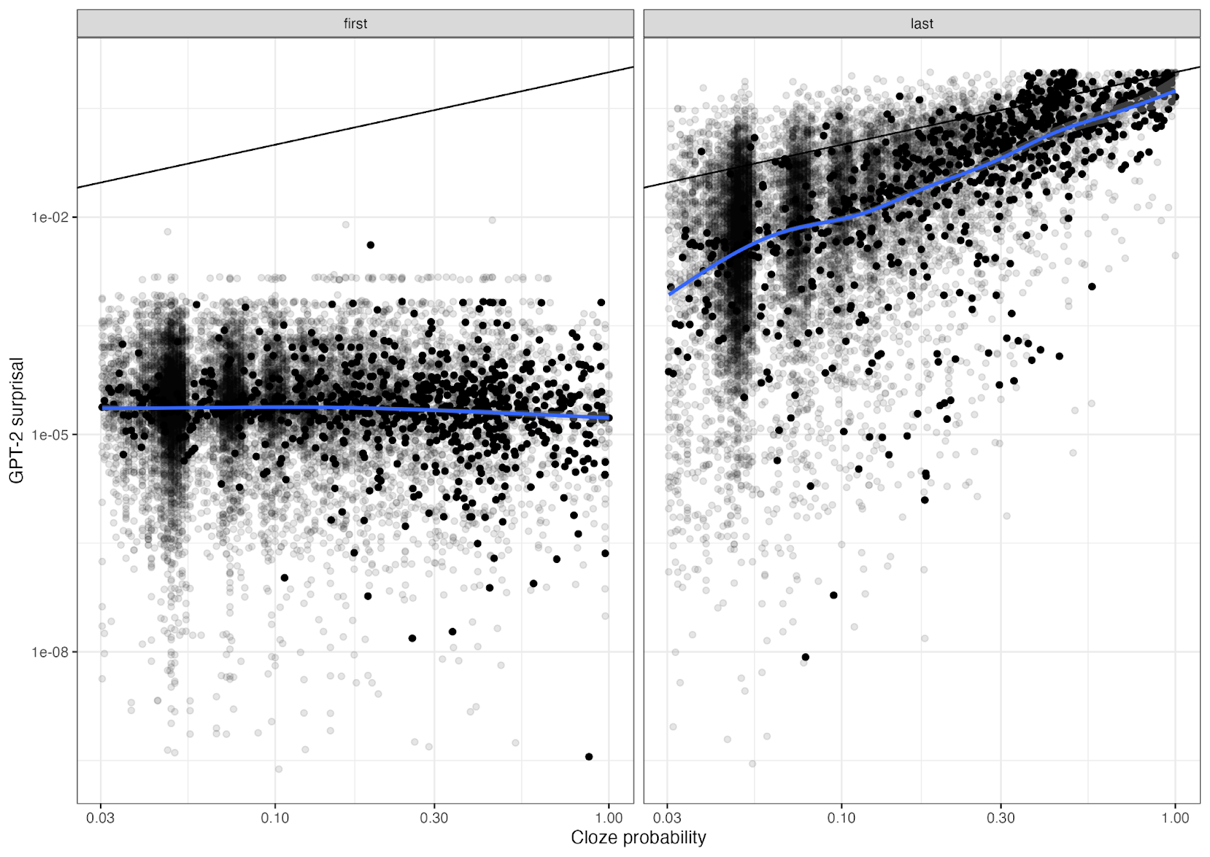}
	\caption{ Demonstration of correlation between language model and human cloze probability in
		\citeauthor{peelle_completion_2020}'s \citeyearpar{peelle_completion_2020} completion norms
		for GPT-2's first and final layers. Dark points represent cloze level-specific means; lighter points represent
		item-specific values. Solid line represents \(y = x\), indicating a perfect correlation. Model
		probabilities are faceted by layer (first on the left vs. last on the right). }\label{fig:gpt2_cloze_correlations}
\end{figure}

The first analysis assesses how the LM probabilities across layers correlate with human cloze
probabilities, then zooms into a comparison between
the first and later layers of GPT-2, which we illustrate in \zcref{fig:layerwise_clozeprob_correlations} and \zcref{fig:gpt2_cloze_correlations}, respectively.

In this first analysis we compute the correlation between \citeauthor{peelle_completion_2020}'s
\citeyearpar{peelle_completion_2020} cloze probabilities and the three LLMs' probabilities.
As expected given the surface similarity between the cloze and LLM prediction task, we obtain high item-level correlations depending on the layer and LLM.
The correlations typically become stronger from layer to layer, with the highest correlations uniformly at the final layers across Pythia, GPT-2, and RoBERTa, suggesting that the layerwise computation occurring within LLMs transforms its representations toward the prediction objective, a pattern we present in \zcref{fig:layerwise_clozeprob_correlations}.
 Spearman’s correlation is the most appropriate measure to use here due to the bounded range of the data, and the strongest correlation to cloze probabilities of \num{0.452} comes from the final layer for Pythia. We observe slightly weaker correlations when measuring Pearson’s correlation at \num{0.425} for the same model and layer. The correlation between the probabilities at the final layer of both Pythia and GPT-2 is extremely high (Pearson's \(ρ\) is \num{.92}).

In spite of the promisingly strong correlation between human cloze responses and LLM probabilities, a closer inspection reveals important points of divergence between human and machine next-word predictions.
The correspondence between human and LM behavior is not homogenous across predictions, such that there is less agreement (greater error) between humans and LMs at lower-probability tokens, possibly partially due to sparsity in the human responses.
A median split analysis demonstrates that while the correlations are stronger at higher cloze probability levels
(\(B_{\mathrm{upper}} = 0.04\), \(t(35456) = 45.78\), \(p < 0.001\)), the item-level correspondences are highly variabl, replicating a prior finding from \citet{jacobs_human_2020}.
\zcref{fig:gpt2_cloze_correlations} illustrates this for GPT-2.
We include the analogous plots for RoBERTa and Pythia in \zref{app1}.

\subsection[Study 1.2]{Study 1.2: Layerwise encoding of next-word probabilities across
models}\label{sec:sudy_layerwise_encoding}

The pattern of results from \ztitleref{sec:study_layerwise_correlation} becomes more interesting
when we consider that the layerwise computation of next-token probabilities is possible because each
layer partially preserves linguistic information from prior layers, but that different language models
likely produce surprisal estimates in very distinct ways. 
This difference between language models is what we aim to investigate in this second analysis.
In \ztitleref{sec:sudy_layerwise_encoding}, we again apply the
logit lens to each layer of RoBERTa, Pythia-160M, and GPT-2 and probe how the specific activations (language model probabilities) of
different lexical predictions emerge in each model.

\begin{figure}
	\centering
	\includegraphics[height=0.33 \textheight]{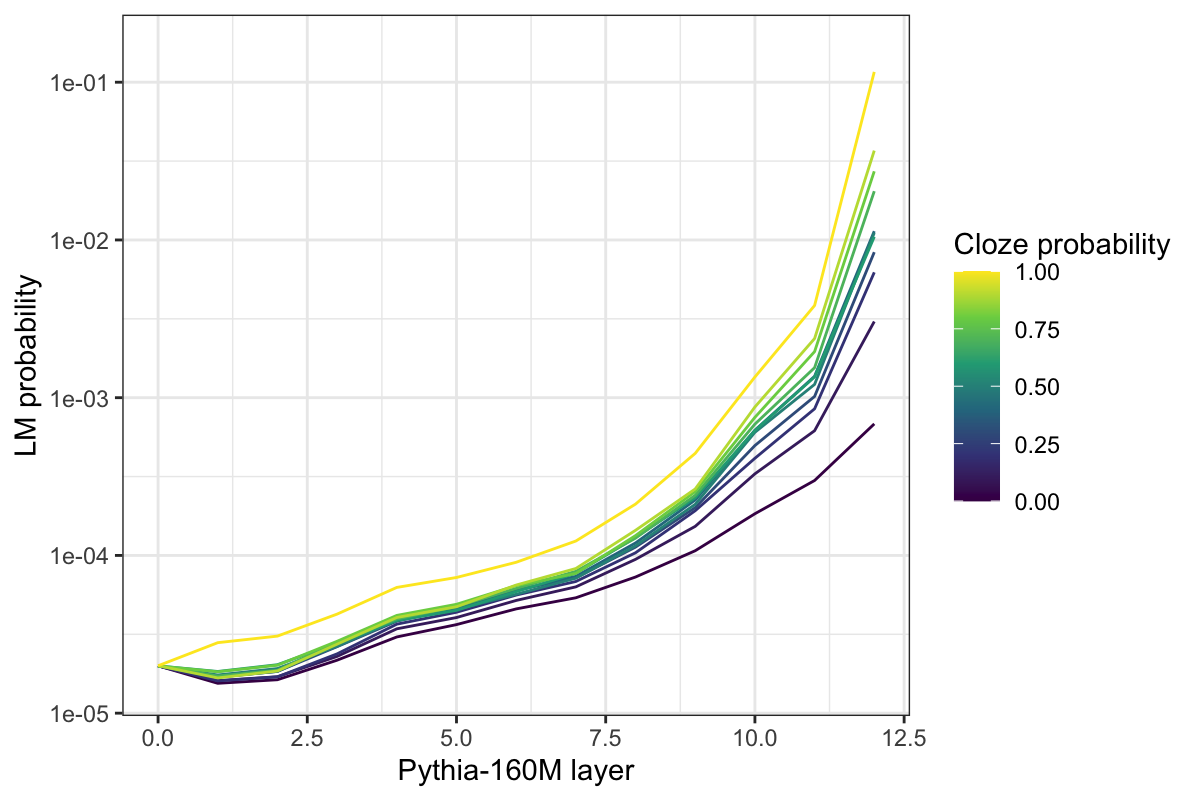}
	\includegraphics[height=0.33\textheight]{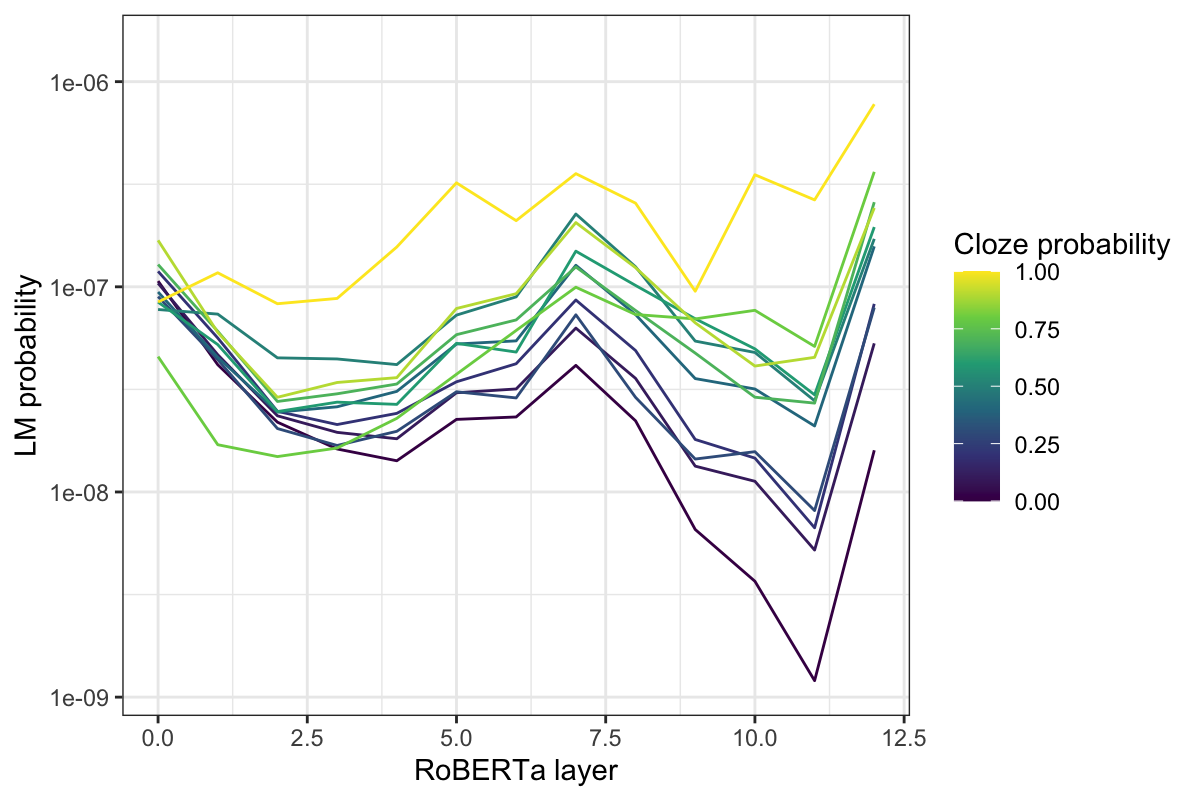}
	\includegraphics[height=0.33\textheight]{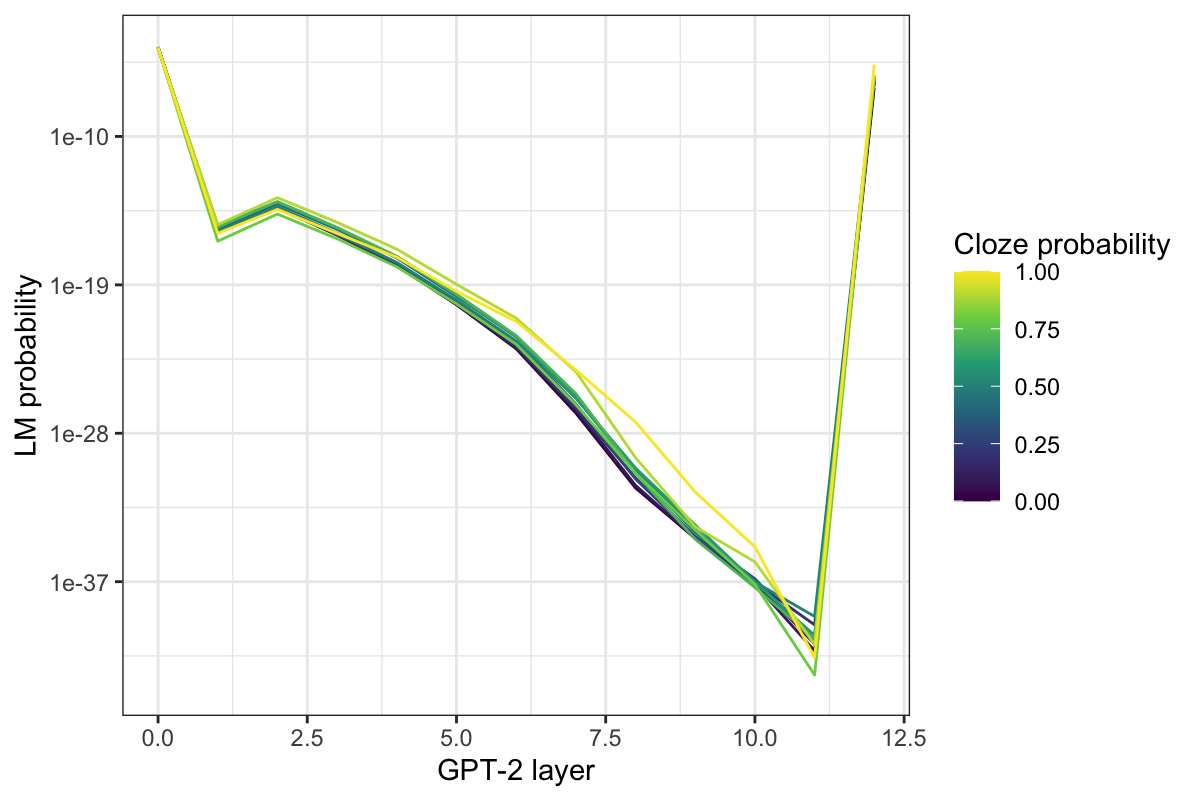}
	\caption{Next-word probabilities through the logit lens across layers and
models.}\label{fig:layerwise_logit_lens}
\end{figure}

In light of the strong correlation between cloze probability and surprisal
(\ztitleref{sec:study_layerwise_correlation}), we visualize the surprisal estimate for each
completion across layers with separate bins for continuations that vary in their human cloze
probability, such that some completions belong to strongly constraining contexts, and others less
so (\zcref{fig:layerwise_logit_lens}).

As shown in \zcref{fig:layerwise_logit_lens}, starting from the very first layer and across all three models, high-probability cloze completions are
typically assigned the highest probabilities and the lowest-probability completions are assigned the
lowest probabilities. 
However, we can see that the models start to significantly diverge when qualitatively examining the trajectory of probabilities that can be assigned to human responses across layers.
Interestingly, both GPT-2 and RoBERTa exhibit a sharp change in the strength of known next words between the penultimate layer and the last hidden state, which is most tuned to the language modeling head, which produces the logit values by transforming the model's encoded representation of the context into a distribution over next-token probabilities \citep{rogers_primer_2020,tenney_bert_2019}.
By contrast, Pythia shows a more gradual change from
layer to layer, with each response gradually increasing in probability across
layers.
This is most likely due to the geometry of Pythia embeddings more generally
\citep{machina_anisotropy_2024,lee_geometric_2025}.
These results are also consistent with previous findings arguing that the language modeling objective changes hidden states most in intermediate layers because different linguistic
structural interactions have been computed by the system
\citep{rogers_primer_2020,tenney_bert_2019}.

\subsection{Discussion}

In \ztitleref{sec:expe_lexical_predictability}, we showed that even though different models can correlate to similar degrees with human data, there are crucial differences between human and models and between models and models.

The divergence in the trajectory of probabilities observed in this experiment is in line with the fact that different LLMs are known
to encode different linguistic properties to different extents across the distinct layers of hidden
states \citep{tenney_bert_2019}.
Thus, far from producing theoretically-equivalent surprisal estimates, different LLMs imply distinct computational and algorithmic commitments.
While this may appear obvious, there are clear theoretical consequences for cognitive modelers.
In particular, the models represent different information in their own computation of surprisal depending on whether they encode context bidirectionally or serially
\citep{devlin_bert_2019,radford_language_2019}.
This in turn instantiates different choices about the nature of the information leveraged in the computation of surprisal.
Additionally, other architectural decisions for the different models similarly constrain the computations that are able to be made \citep{machina_anisotropy_2024}, how information flows across layers, and the downstream surprisal estimates that result.

Additionally, the results of \ztitleref{sec:expe_lexical_predictability} demonstrate that the specific implementations of the computation of next-word probabilities are clearly highly diverse within computational and human systems.
In this sense, merely focusing on the  magnitude of the correlation between humans and language models can be misleading for cognitive modelers.
While some language model behaviors correlate better than others with human responses in the cloze
task, the lack of an exact reason for any correspondence between the two should raise concerns. To
take one example, the next-word prediction models (GPT-2 and Pythia) both correlate with cloze
probabilities more strongly than the masked language model (RoBERTa). One could argue on the basis
of predictive powers (e.g., \(Δ\textrm{LL}\) or \(R²\)) that this is evidence that next-word prediction is a better model
of sentence comprehension in humans and machines \citep{michaelov_strong_2024}.
Indeed, many computational psycholinguists have largely moved on from using masked language models in favor of next-word prediction systems.
However, the theoretical backing for these decisions has largely not been provided, and instead models are adopted to the extent that they produce the greatest improvements in accounting for variation in reading times.
Any number of decisions that shaped the development of GPT-2 or Pythia or RoBERTa -- from data to model architecture -- could explain why a model is better at matching human performance on empirical grounds. 
Even models of different sizes may vary in their capacity to match human language processing \citep{10.1162/tacl_a_00548} but the causal reasons for these patterns are difficult to infer.
Regardless, greater PP does not necessarily justify a model's use, especially when the computations are
known to differ in important ways \citep{cummins_how_2000}.

\section[Experiment 2]{Experiment 2: Lexicality of inner computations of large language
models}\label{sec:expe_lexicality}

In \ztitleref{sec:expe_lexicality}, we present another means to assess the behavior of these three
models. We demonstrate that each model represents a different linking hypothesis about how
next-word predictions are computed. Of particular interest is the representational content of
the hidden states within the language models, particularly how much the states encode the model's
beliefs about future states as \emph{words}, as opposed to other linguistic objects (e.g., semantic or syntactic categories or abstractions; \citealp{jacobs_towards_2025}). Such an analysis is possible
because layerwise transformations all work toward the eventual goal of producing a vector of lexical
activations, and Transformer architectures propagate different transformations of lower layers up the network \citep{barenholtz2026trajectorydynamicslanguagemodel,domenichelli2026linguistic}.

\subsection{Data}

We probe the process of composition in each model by analyzing the ten top-ranked responses after
applying the \enquote{lexical lens}. The lexical lens allows us to examine model outputs directly, in a human-interpretable way, by analyzing the top (sub)words for each logit lens. In this experiment, we deploy the Provo Corpus, which contains \num{2398} sentence preambles taken from naturalistic web data sources. It contains just over \num{41000} unique preamble-cloze response pairs \citep{luke_limits_2016,luke_provo_2018}. The Provo Corpus is widely cited in computational psycholinguistic analyses and has been instrumental in arguments illustrating the existence of a correlation between language model surprisal and reading times \citetext{\citealp{shain_large-scale_2024}, a.o.}.

\subsection{Method}

To summarize the output of each model's predictions, we propose a metric that estimates the
\enquote{wordlikeness} of a language model's top predictions, which we call \emph{lexicality}.
Formally, lexicality is computed as the proportion of the top $k$ next-word guesses by the language
models that correspond to lexical items in some lexicon \citetext{here, SUBTLEXUS; \citealp{brysbaert_moving_2009}}. To the extent that Pythia, GPT-2, and RoBERTa are computing
dissimilar representations from each other, such dissimilarities may be evident in the lexicality of
word guesses across each layer of the model. For illustration, we present a
representative lexicality output from the three models under consideration in \zcref{tab:model_completions}.

\begin{table}[t]
	\centering
	\caption{ Top \num{10} LLM completions of \citeauthor{luke_limits_2016}'s
		\citeyearpar{luke_limits_2016,luke_provo_2018} sentence fragment \enquote{When early Europeans
		discovered Easter Island, its somewhat isolated ecosystem was suffering from the effects of
		\_\_\_}, across models. For the sake of brevity, we only include an illustrative sample of layers.}\label{tab:model_completions}
	\bigskip
	\begin{tabular}{cccccc}
		\toprule
		\multicolumn{2}{c}{\textbf{Pythia}} & \multicolumn{2}{c}{\textbf{RoBERTa}} & \multicolumn{2}{c}{\textbf{GPT-2}}\\
		\cmidrule(r){1-2}\cmidrule(lr){3-4}\cmidrule(l){5-6}
		{2} & {13} & {8} & {13} & {1} & {9} \\
		\midrule
		\}\}\})\$ & the & . & deforestation & the & the \\
		\textbackslash"\}**). & a & fragmentation & pollution & a & a \\
		course & climate & . & erosion & \textbackslash" & an \\
		\}\}\}(\textbackslash\textbackslash{} & an & growth & drought & , &
		\textbackslash" \\
		\}\}\}\{\textbackslash\textbackslash{} & disease &
		\textless/s\textgreater{} & disease & \textquotesingle{} & this \\
		\}\}\}(\{\textbackslash\textbackslash{} & its & contamination &
		isolation & an & , \\
		\textbackslash"\}{]}(\# & cold & colonialism & flooding & in & its \\
		\^{}{]}(\# & over & events & man & - & their \\
		))**(- & global & development & hurricanes & this & in \\
		\}\}\}\_\{\textbackslash\textbackslash{} & starvation & depression &
		humans & & all \\
		\bottomrule
	\end{tabular}
\end{table}

\begin{figure}
	\centering
	\includegraphics[width=\textwidth]{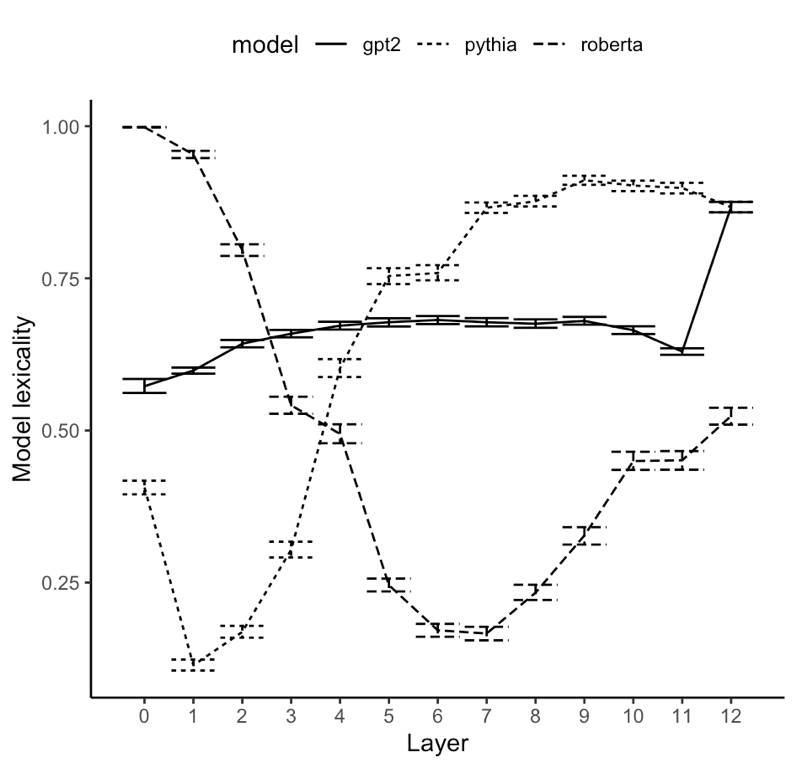}
	\caption{Layerwise lexicality changes for \(k=10\) over the \num{2398} preambles of the Provo
Corpus (Luke \& Christianson, 2016, 2018).}\label{fig:layerwise_lexicality}
\end{figure}

A visual inspection of the average lexicality across all preambles, for each layer and model, suggests that different layers reveal increasingly contextual, linguistically-interpretable predictions, consistent with early Transformer interpretability analyses \citep{tenney_bert_2019},
but with major layerwise differences across models (\zcref{fig:layerwise_lexicality}). Focusing on the shapes of these curves, we see massive fluctuations in lexicality across layers and strikingly different patterns between models.
Pythia begins with very low degrees of lexicality that rapidly rises to high levels, where it plateaus around layer \num{10} of \num{13}.
We believe that this is a consequence of differences between the embedding and unembedding matrices (as discussed in \zcref{sec:models}). These results
also help to explain the very low probabilities assigned to particular tokens; strings that are
interpretable as words (i.e., human predictions) are simply not available to Pythia at early stages.
By contrast, RoBERTa shows a distinct cup-shaped pattern, beginning at high degrees of lexicality that slowly
morph by the middle layers (\numrange{6}{8}) into delexicalized subwords.
An inspection of RoBERTa’s top predictions in these interim layers reveals rare character sequences or sequences of punctuation symbols.
GPT-2 shows yet a third pattern, such that it maintains a relatively high
degree of lexicality throughout, rising somewhat between the first and antepenultimate layer, before
dropping at the layer before the last and rising sharply at the output layer. In conjunction with
\zcref{tab:model_completions}, we can see that pre-final layers of GPT-2 most strongly
\enquote{encode} highly frequent individual words. At the last layer, GPT-2 rises to Pythia's levels
of lexicality to predict more word-like tokens that linguistically resemble Pythia's.

\subsection{Discussion}

\ztitleref{sec:expe_lexicality} further builds on the findings of
\ztitleref{sec:expe_lexical_predictability}, and demonstrates that the internal computations of
Pythia, GPT-2, and RoBERTa possess vastly different geometries and linguistic trajectories.
A lexical lens functions in a human-interpretable way to provide complementary evidence to 
\ztitleref{sec:expe_lexical_predictability}, which showed that the production of probabilities of
potential continuations changes in starkly different ways across layers.
\ztitleref{sec:expe_lexicality} shows that when not considering human responses, but rather even the
models' own top-ranked guesses, the way that the model arrives at these guesses similarly is the result of
vastly different computations. While these results are not particularly surprising, as the
engineering advantages of different types of language models for different types of tasks has been
well-established, the results underscore the challenge in applying LLMs as a tool to test Surprisal
Theory, as the models are plainly not interchangeable in the way that they arrive at their
predictions.
We turn to these issues in the General Discussion.

\section{General Discussion}

A growing literature has suggested that at least under some circumstances, readers engage in
linguistic prediction at the level of specific phonological forms, individual words, and
higher-order structures such as syntax
\citep{haeuser_predictive_2024,luke_limits_2016,willems_prediction_2016}.
Theories that argue that language processing essentially consists of predicting upcoming linguistic content have drawn analogies between the objectives that language models are designed to solve (e.g., next-token prediction) and the process of anticipating potential next words or structures.
As such, correlations between these two processes have been taken as evidence for Surprisal Theory \citep{michaelov_strong_2024,shain_large-scale_2024}.

Our experiments illustrate how even though the surprisal values computed by different LLMs can appear superficially similar, they encode different types of information and are computed by each model in strikingly distinct ways.
The logit lens revealed that the probabilities of likely next words fluctuate wildly across different layers of different models.
The lexical lens provides transparency about the very different patterns for the three models as they generate their predictions.
We observed that model behavior varies across architectures and training objectives. These results underscore how the generation of a distribution of predictions can arise by processes unique to different model architectures and training objectives.
On the basis of a correlation to human cloze probabilities, we might conclude that Pythia is the most human-like of the models we analyzed, but it is clear that the way it arrives at its predictions is distinct from RoBERTa and GPT-2.
Thus, it is critical to evaluate the particular commitments made by each model about the relationship between input and output, which shapes the computations performed by the system.
If different LLMs change the linking hypothesis between probabilities and effort, then we are obfuscating what once made Surprisal Theory an informative hypothesis to psycholinguists.
Researchers using LLM-derived surprisal thus end up accidentally operationalizing Surprisal using different underlying mechanics.

We want to now cast a critical lens on how the wide-spread use of LLM probabilities in computational psycholinguistics has influenced theories of human language processing.
In the remainder of this paper, we discuss what we believe are critical issues with the use and interpretation of LLM-based surprisal in modern psycholinguistics.
We ask whether our understanding of Surprisal Theory as a computational-level account of language processing has improved in light of the ever-growing number of works that correlate surprisal with psychometric indices of cognitive processing difficulty.
We then examine the extent to which we can claim that using LLMs is “representation-agnostic” to providing evidence for or against Surprisal Theory \citep{futrell_lossycontext_2020}.
Have these findings improved our theories of linguistic cognition in general? 
In other words we ask: \enquote{\emph{What kind of theory does surprisal -- the measure -- encode, and what is it meant to evaluate?}}

\subsection{Claim: surprisal is not a computational-level proxy}

To better understand the issue we are trying to raise here, let us ask what exactly it means for
something like Surprisal Theory to be interpreted as a computational-level theory in Marr's sense.
Marr's approach has served as a valuable guide for theorizing within the cognitive sciences, since
it offers a way to conceptualize the same object (e.g., a cognitive system or capacity) at different
degrees of abstraction.\footnote{ Though we should be wary of committing too strongly to a strict
separation between these levels
\citep{griffiths_rational_2015,poggio_levels_2012,pylyshyn_computation_1984,embick_towards_2015}.}

Traditionally, theories at Marr's Computational level are meant to characterize the goal of the computations performed by a system, why it is appropriate for the system under study, and what the logic of the strategy for carrying out such computation is \citet{marr_vision_1982}.
Marr's levels call for researchers to be explicit about the formal commitments and
demands of their theories  \citep{guest_how_2021}, and thus offer a good framework for leveraging
computational modeling work in theory building. %

For this perspective to be insightful for cognitive theories, computational-level linking hypotheses should by definition provide transparent, interpretable connections
between observable measures of behavior and hypotheses about the unobservable mechanisms of the
underlying system \citep{spivey_linking_2025,tanenhaus_-line_2004}. 
That is, a computational-level perspective should allow researchers to understand the function that is computed by the system enough to be able to discern its fundamental properties, and how they abstract from its many possible algorithmic realizations.\
Surprisal Theory is therefore an abstraction that characterizes the nature of a relationship, but does not need to be implemented by a particular system.

However, researchers who extract surprisal values from particular models on the basis of their fit to human behavioral data, or because they are perceived as generating the most humanlike language, have selected a particular algorithmic implementation.
Understanding the content encoded by the computational transformations that take place between the input and output of language models should be a minimum prerequisite for deciding whether a particular system can be used even as a computational-level characterization of language processing.
Given that LLMs are essentially black boxes \citetext{see \citealp{berkeley_curious_2019}, for a clear-eyed review}, their use inhibits researchers' ability to identify such abilities.
In this sense, even if truly representation-agnostic estimates of surprisal were possible, it is not clear why they would be desirable from an epistemological perspective: simply estimating a quantity without committing to
hypotheses about the nature and structure of what is being computed is of very little usefulness for
both theory crafting and validation.
LLM-derived surprisal is therefore difficult to integrate into a computational-level description.

\subsection{Claim: LLM surprisal Lacks Cognitive Commitments}
At the time it was first introduced, Surprisal Theory was a captivating theoretical advance that
combined the strengths of probabilistic constraint satisfaction approaches
\citep{seidenberg_probabilistic_1999} with symbolic reasoning over syntactic
categories or structures
\citep{boston_parsing_2008,demberg_data_2008,frazier_making_1982,frazier_syntactic_1985,jelinek_computation_1991,kaplan_augmented_1971,rauzy_robustness_2012,yngve_model_1960}.
In line with Marr's original formulation, \citet{hale_informationtheoretical_2016} explicitly calls
for a computational-level theory of sentence processing as a high-level description of what a system
is doing, assuming that \textquote{Surprisal \textelp{} specif\textins{ie}s what the difficulty
level of parsing a word should be in terms of structures defined in a language model.} Hale
acknowledges that the next fundamental step in building explanatory linking hypotheses is to then
provide a plausible model of the underlying process, in terms of compatible mechanisms (e.g.,
chunking).

The dual commitments to the characterization of (1) representations and (2) mechanisms underlying
language processing are conspicuously \emph{absent} in some recent research. Consider, for example,
works focusing on the shape of the correlation between surprisal and comprehension difficulty. 
Whereas \citet{hale_probabilistic_2001} proposed that a word's surprisal should be monotonically related to processing difficulty, \citet{smith_effect_2013} claimed that the relationship should be log-linear in nature.
Recent analyses using massive language models have corroborated these earlier claims, whose evidence relied on correlations between reading times and non-neural statistical language models \citep{shain_large-scale_2024,smith_effect_2013}. 
However, others have argued that a word's predictability is linearly correlated with its processing difficulty \citep{brothers_word_2021}.
As a compromise position, 
\citet{szewczyk_context-based_2022} argued that a combination of linear and log-linear components is
necessary to account for the full pattern of neurobehavioral data. 

In the conduct of these
studies, many researchers have tested hundreds of varieties of language models that possess
different architectures and training objectives, to hopefully identify linear, log-linear, or other
types of relationships between human data and LLM estimates of next word probabilities
\citep{oh_leading_2024,shain_large-scale_2024,timkey_eye_2025}. Researchers in this area offer
minimal hypotheses about the merits of different models or representations from a cognitive
point-of-view. That is, the level of description at which surprisal is formulated in recent work
falls short of the specificity that Marr called for in characterizing the properties of the function
that is being computed, as well as of the insights original work on Surprisal Theory claimed we
could get from pursuing this research enterprise.

\subsection{Claim: LLM surprisal is not a monolithic construct}
Another challenge with this approach is the generality of the assumed importance of language model
probabilities for many aspects of human language processing. Commonly, researchers argue for the
importance or reflection of surprisal on human cognition based on improved correlations in
regression models of psychophysical data. In particular, researchers seem to treat different Dependent Variables (DVs) as
effectively interchangeable and broadly reflective of the same cognitive process \citetext{e.g.,
probabilistic pre-activation of some upcoming linguistic category;
\citealp{campanelli_modulatory_2018,li_decomposition_2023,shain_large-scale_2024}}. For example,
many studies have compared and contrasted a surprisal-based N400 response to a similar relationship between surprisal and
reading times, though these two measures are very likely generated by different cognitive mechanisms
\citetext{\citealp{frank_word_2013,michaelov_strong_2024}; a.o.}.

While a tendency to use psychometrics as interchangeable proxies for cognitive processes is common
in psycholinguistics, there has nonetheless been some care in distinguishing general indexes of
\enquote{effort} (e.g., reaction times, fixations, pupil dilation) to more refined indexes of mental operations
sensitive to different processes and levels of representations
\citetext{\citealp{stanojevic_modeling_2023,timkey_eye_2025,DeSanto2025capturing}; a.o.}. Work applying
LLM-generated surprisal to human data as a blanket measure loses track of these distinctions. It
confuses potential explanations of the link between surprisal and effort with more theoretically
powerful explanations of the link between predictability as a property of language and the nature of
predictive mechanisms.

\subsection{Claim: LLM surprisal Obfuscates Theoretical Questions}
Perhaps even more problematically, this focus on predictive power (PP) \emph{per se}, as well as the practice of data-fitting and making theoretical conclusions on the basis predictive power, has been fundamentally diminishing the field’s ability to critically incorporate computational results into our broader theory-building enterprise.
For example, analyses of PP have been used to argue that cloze probabilities like those we analyze above are poor estimates of human linguistic prediction, and that language model probabilities are more valid estimates of the type of linguistic knowledge that readers use.
Researchers who have focused on selecting the best surprisal estimate for their analyses see coarser measurements of predictability relative to LLM logits as a shortcoming
\citep{smith_cloze_2011,smith_effect_2013,szewczyk_context-based_2022}, since answers consist of one
or a few discrete words \citep{taylor_cloze_1953}.
This is despite the fact that cloze probabilities have long been used as a proxy for the predictability of words in texts
\citep{de_varda_cloze_2023,federmeier_multiple_2007,luke_limits_2016}, are a
reliable predictor of reading times and neural signals \citep{szewczyk_context-based_2022} and are generated from humans engaged in the processes our cognitive theories attempt to explain.
Similarly, on the basis of PP, researchers have argued that predicting words in the cloze task does
not engage the same prediction mechanisms that are thought to support efficient language processing
\citep{smith_cloze_2011}, and that cloze probabilities should be abandoned as proxies for
predictability.
In the absence of a cognitive explanation, however, differences in PP are not sufficient support for this argument.

The focus on predictive power of surprisal has also obscured recent issues that surfaced with the Natural Stories Corpus \citep{futrell_natural_2018}, an
annotated self-paced reading (SPR) dataset that is incredibly popular in (computational)
psycholinguistics research exploring LLM-generated surprisal correlation to human effort in reading
\citetext{\citealp{shain_large-scale_2024,10.1162/tacl_a_00548}; a.o}. As of this writing, the corpus has been
cited in over \num{120} papers, according to \href{https://web.archive.org/web/20251129175647/https:/www.semanticscholar.org/paper/The-Natural-Stories-corpus\%3A-a-reading-time-corpus-Futrell-Gibson/056366476e728476a767d126ae5881aa5b8468a2}{Semantic Scholar}.
 These issues derive from several fundamental errors in how the corpus was compiled, with SPR data in the original release misaligned by one position.\footnote{The
\href{https://web.archive.org/web/20251031181830/https:/github.com/languageMIT/naturalstories/blob/4700daad696e942f5aba23c957a7423d0de66612/README.md}{official
repository} states: \textquote{Update 2025-05-12 We have discovered that the SPR data in the
original release was misaligned by one position. \textelp{} If you have been using the dataset, we
recommend re-running your analyses on the realigned data.}}
The most visible consequence of
recognizing this error in Natural Stories has been a recommendation to \enquote{rerun past analyses}.
The implicit assumption here seems to be that as long as the results of our data-fitting processes are unchanged (e.g., based on statistical comparisons), then the error was inconsequential and we should not worry.
However, given the success of previously-published surprisal measures on this
corpus, and the (implicit and explicit) assumption that surprisal should capture some measure of effort at the current word, such a misalignment error should have led researchers to reconsider why
exactly their surprisal analyses had been so successful. In other words, if a researcher believed
they were modeling processing effort at word \(w_i\), but in reality they were modeling \(w_{i+1}\)
(or \(w_{i-1}\)), what does that say about the theories of cognitive processes they have been
building based on that correlation?
At the very least, errors of this magnitude should force us to re-evaluate what we thought we understood about what a measure or model means. If we are interested in building informative theories of cognitive systems, and we think our computational constructs should represent interpretable linking hypotheses between behavior and the cognitive processes under study, we should ask what we are actually learning from a metric that does not seem to be sensitive to such a core element of the process we are trying to understand (in this case, \emph{which} word we are measuring effort at).

\subsection{Claim: LLM surprisal Obfuscates Representational Assumptions}
The problem is also evident when examining the objects that LLMs predict in the first place.
Subwords are nearly
always the object of prediction in these models
\citep{nair_words_2023,oh_surprisal_2021,oh_leading_2024}. That is, a model may not normally be
predicting words at all, but rather character sequences that sometimes correspond to words.
\citet{oh_leading_2024} rightly point out that in experiments where the written forms of words are separated by spaces, human beings read words, and therefore advocate (on the basis of PP) for computing probabilities that are aware of word boundaries and character positions.
While this advice is seemingly innocuous, incorporating word boundary information is a fundamental change to the computation of surprisal.
Still others have argued that computing the probabilities of character sequences is more cognitively consistent with decades of research into visual word recognition \citep{apel_orthographic_2019,vieira_language_2025,oh_surprisal_2021,oh_leading_2024}.
However, these methods are impractical to compute and difficult to incorporate into most analysis pipelines because computing the conditional probability of an entire character sequence is inefficient.

More relevant for the cognitive modeler, these two \enquote{solutions} both represent inductive biases about the nature of words, about the relevant linguistic levels at which readers are computing predictions, and yet the cognitive algorithm that ties surprisal to the allocation of attention is assumed to be unchanged.
Arguing for representational changes on the basis of PP directly contradicts the proposal that surprisal is a tool to evaluate a computational level description of the same system, when the information involved in the characterization(s) of such system differs in several fundamental aspects \citetext{e.g., characters versus subwords versus words;
\citealp{oh_surprisal_2021,oh_leading_2024}}.

\subsection{Claim: LLM surprisal Reduces Behavior to a Benchmark}
Relatedly, the recent pursuit of \enquote{better} LLMs for modeling different DVs \citetext{e.g., \citealp{oh-2026-model}} has re-cast analyses of human behavior into a problem for which it is possible to achieve something like \enquote{state-of-the-art} performance.
Under this approach, this reasoning states that better models will attain a closer fit to a DV on the basis of \(Δ\mathrm{LL}\) or \(R²\), such that finding a significant contribution to model fit is seen as evidence of that variable's importance to
psycholinguistic processing.
From this fitting-centric perspective, the way to improve large language models as a model of human cognition requires us to make them match human beings.
For example, \citet{oh-2026-model} highlight that a major discrepancy between human and LLM processing lies in the greater memorization capacity of LLMs and that limiting memory
capabilities of language models is critical for improving fit to human data \citetext{e.g., by applying lossy context surprisal; \citealp{futrell_lossycontext_2020}}.
\citet{oh-2026-model} argue that LLMs should be adjusted to be more human-like and to possess more limited memory representations.
However, if the researcher alters the memories of the model to be susceptible to forgetting or
illusions, this changes the model that computes surprisal itself, and this approach changes the
algorithm in the first stage of the allocation of attention. This call to match LLMs to humans is a
reversal of the philosophical approach to modeling cognition from first principles and embodies the
split between classical and modern connectionism described by \citet{guest_metatheory_2025}.

The conclusion that we should simply alter LLMs to behave in a more human-like way reveals the
scientific dead-endedness of the current enterprise. Many researchers in the field continue to chase
variants of a statistical measure to continue maximizing predictive power, with little reflection on
whether this ceiling has already been reached and whether it would yield useful information. That
is, having developed estimates of language statistics that apparently go beyond what humans can
track (and therefore result in decreasing predictive power for human behaviors), the conclusion is
\emph{neither} that surprisal is only useful or predictive up to a point, \emph{nor} that we need to
reconsider the cognitive processes and constraints that would result in a \enquote{sub-optimally estimated} surprisal being the better predictor for human behavior.

\subsection{In Sum: surprisal is Not an Explanation}
In a sense then, treating LLM-surprisal as a good, generally valid implementation of a vaguely specified version of \enquote{Surprisal Theory} gives us neither a good computational-level theory of predictive processes in language, nor an algorithmic-level theory detailed enough to ask cognitively relevant questions about predictive mechanisms \citep{culbertson2026language, murphy2026machine, resnik2026models}.
In different terms, we end up without a good or useful computational-level account because the opaqueness of the models used to estimate the surprisal \enquote{measure} prevents us from inferring anything about the aspects of the system that computational-level descriptions should help us understand (e.g., core properties of the function, restrictions on the domain of the function, the impact of representational choices, etc.).
Furthermore, we end up lacking a good algorithmic-level account because most practitioners in this line of work don’t manipulate the algorithmic details explicitly enough to be informed by whether they matter or not for the generalizability of the theory.

\subsection{Where Do We Go From Here?}

In this work, we want to advance the perspective that the major goal of applying statistical language modeling (broadly construed) to psycholinguistics should not be to establish a correlation or even to test which measures best fit the behavioral data.\ 
Rather, any theory of sentence processing
should provide an account of what knowledge we use when comprehending language
\citep{embick_towards_2015,guest_how_2021,slaats_whats_2025,staub_predictability_2025}. 

We thus wish to offer suggestions on how to fruitfully investigate Surprisal Theory as an informative account of language processing.
First, the multiple realizability problem ensures that surface-similar surprisal values
will arise from many different types of language models. While multiple-realizability is intrinsic
to computational-level theories, a computational-level account is informative as a characterization
of a cognitive system only when it allows us to determine properties of the system that are
independent of the algorithmic specification.
As we argued above, this should make practitioners
cautious of using surprisal metrics extracted from opaque models as interchangeable, and researchers
serving as reviewers should refrain from uncritically requesting adding surprisal (from the latest LLM) as a covariate into analyses of psychophysical data.
For similar reasons, psycholinguists should be careful in uncritically treating surprisal estimates from  dozens or even hundreds of language models as alternative implementations of the same underlying linking hypothesis, and selecting \enquote{the best model} purely on the basis of predictive power.
Instead, researchers should carefully consider how the numerous ways in which these models differ from each other impact the theoretical assumptions about what surprisal is estimating \citep{oralova_surprisal_2025,timkey_eye_2025}.
When extracting surprisal values from a language model, researchers should explicitly commit to which models/architectures they consider cognitively plausible \citetext{cf. \citealp{baggio_plausibility_2024}} and which aspect of a model they believe is necessary to answer their research questions.
Laying out one's hypotheses about the cognitive abilities that allows us to process language successfully,  how they are related to each other, and how they are specified by the models we use would improve our ability to test
different representational and computational assumptions, and help build better theories.

Second, we encourage researchers to move beyond predictive power in justifying their modeling
decisions, and to value cognitively motivated, explainable models over \enquote{state-of-the-art} LLMs.
In this paper we showed that even with effectively identical fits to human data, the computations that
generated surprisal values relied on vastly different representations.
Even selecting between encoder and decoder-type language models requires a theoretical justification describing how human language processing is expected to unfold.
With greater engagement in these deep theoretical issues, it will be possible to bring Surprisal Theory closer to \citeauthor{hale_informationtheoretical_2016}'s
\citeyearpar{hale_informationtheoretical_2016} proposal that researchers should lay out the precise
representations and computations that drive how people's reading times (or neural signals, or eye
movements) change depending on the predictability of some linguistic content.\ 
We believe that Surprisal Theory can
move beyond fitting behavioral data, and support researchers' ability to identify the analogous
components that enable a particular model to solve a cognitive problem.
It is only through this
commitment to explanatory theory building that computational psycholinguistics will arrive at an
understanding of human language processing.

\section*{Acknowledgments}

\noindent The authors would like to thank Grusha Prasad and the three anonymous reviewers for their valuable feedback on previous versions on this manuscript.

\bibliographystyle{elsarticle-harv} 
\bibliography{biblio}

\newpage
\appendix
\section{}\zlabel{app1}

\begin{figure}[th]
	\centering
	\includegraphics[width=\textwidth]{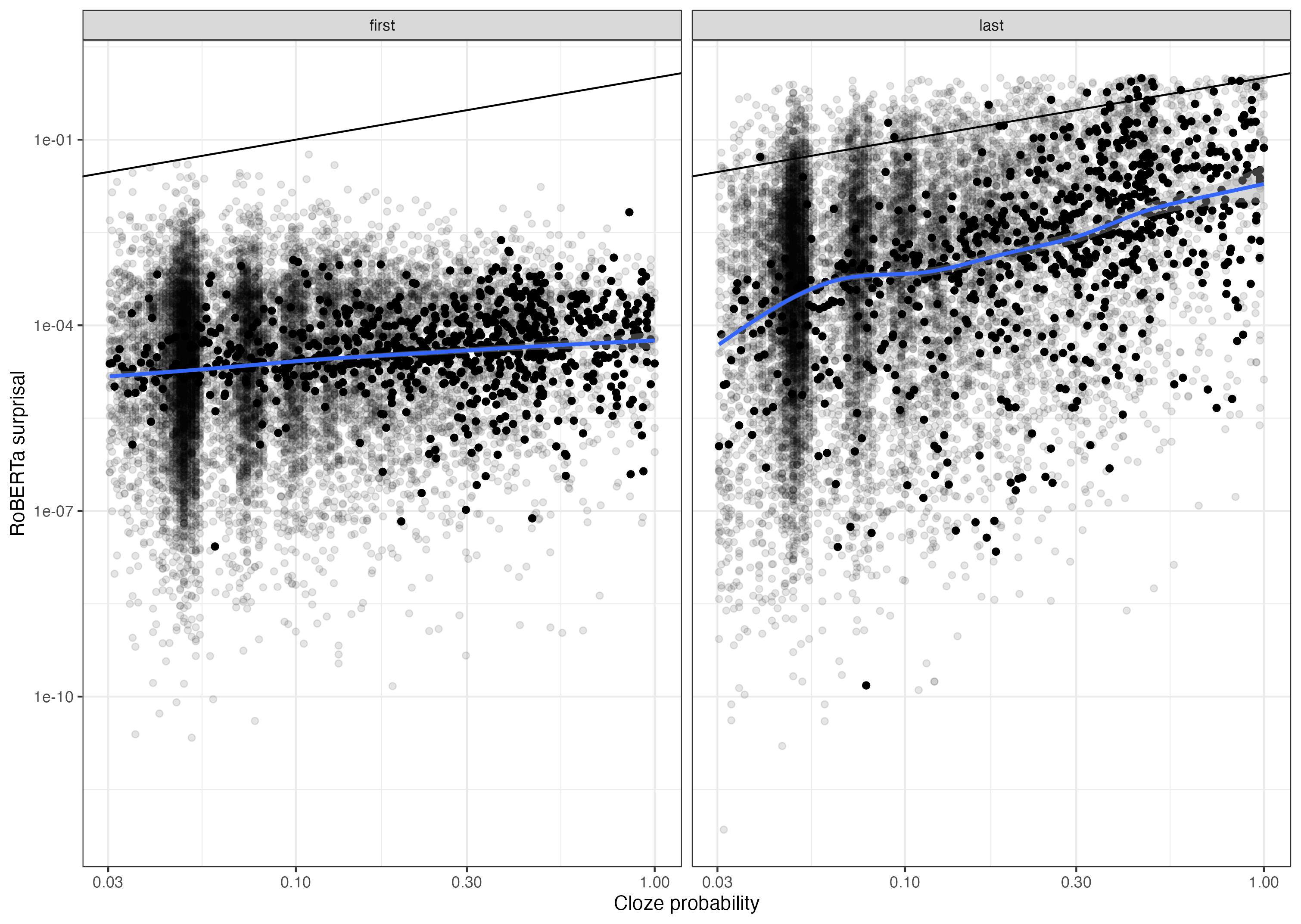}
	\caption{ Demonstration of correlation between language model and human cloze probability in
		\citeauthor{peelle_completion_2020}'s \citeyearpar{peelle_completion_2020} completion norms
		for RoBERTa. Dark points represent cloze level-specific means; lighter points represent
		item-specific values. Solid line represents \(y = x\), indicating a perfect correlation. Model
		probabilities are faceted by layer (first vs. last). }\label{fig:roberta_cloze_correlations}
\end{figure}

\begin{figure}[th]
	\centering
	\includegraphics[width=\textwidth]{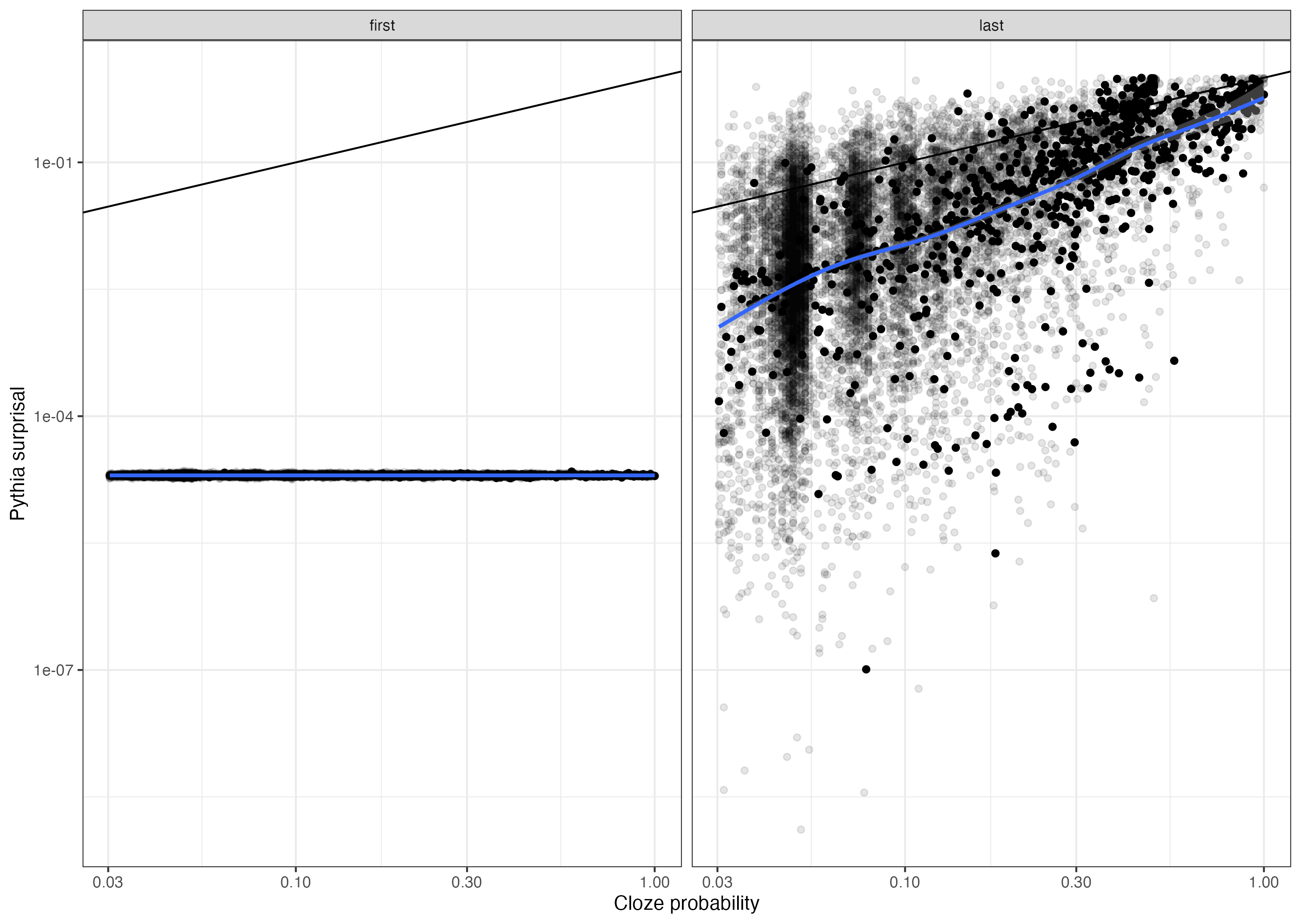}
	\caption{ Demonstration of correlation between language model and human cloze probability in
		\citeauthor{peelle_completion_2020}'s \citeyearpar{peelle_completion_2020} completion norms
		for Pythia-160M. Dark points represent cloze level-specific means; lighter points represent
		item-specific values. Solid line represents \(y = x\), indicating a perfect correlation. Model
		probabilities are faceted by layer (first vs. last). }\label{fig:pythia_cloze_correlations}
\end{figure}

\end{document}